\documentclass[sigconf]{acmart}
\usepackage{algorithm}
\usepackage{algpseudocode}

\AtBeginDocument{%
  }

\setcopyright{acmlicensed}
\copyrightyear{2018}
\acmYear{2026}
\acmDOI{XXXXXXX.XXXXXXX}
\acmConference[MHV '26]{5th Mile-High Video Conference}{February 2–5,
  2026}{Denver, CO, USA}
\acmISBN{978-1-4503-XXXX-X/2018/06}

\begin{document}

\title{Emerging Standards for Machine-to-Machine Video Coding}


\author{
  Md Eimran Hossain Eimon,
  Velibor Adzic,
  Hari Kalva
  and Borko Furht
}

\affiliation{\institution{Florida Atlantic University} \city{Boca Raton} \state{Florida} \country{USA} }

\renewcommand{\shortauthors}{Eimon et al.}


\begin{abstract}
Machines are increasingly becoming the primary consumers of visual data, yet most deployments of machine-to-machine systems still rely on remote inference where pixel-based video is streamed using codecs optimized for human perception. Consequently, this paradigm is bandwidth intensive, scales poorly, and exposes raw images to third parties. Recent efforts in the Moving Picture Experts Group (MPEG) redesigned the pipeline for machine-to-machine communication: Video Coding for Machines (VCM) is designed to apply task-aware coding tools in the pixel domain, and Feature Coding for Machines (FCM) is designed to compress intermediate neural features to reduce bitrate, preserve privacy, and support compute offload. Experiments show that FCM is capable of maintaining accuracy close to edge inference while significantly reducing bitrate. Additional analysis of H.26X codecs used as inner codecs in FCM reveals that H.265/High Efficiency Video Coding (HEVC) and H.266/Versatile Video Coding (VVC) achieve almost identical machine task performance, with an average BD-Rate increase of 1.39\% when VVC is replaced with HEVC. In contrast, H.264/Advanced Video Coding (AVC) yields an average BD-Rate increase of 32.28\% compared to VVC. However, for the tracking task, the impact of codec choice is minimal, with HEVC outperforming VVC and achieving BD Rate of $-1.81\%$ and $8.79\%$ for AVC, indicating that existing hardware for already deployed codecs can support machine-to-machine communication without degrading performance.
\end{abstract}


\begin{teaserfigure}
  \includegraphics[width=\textwidth]{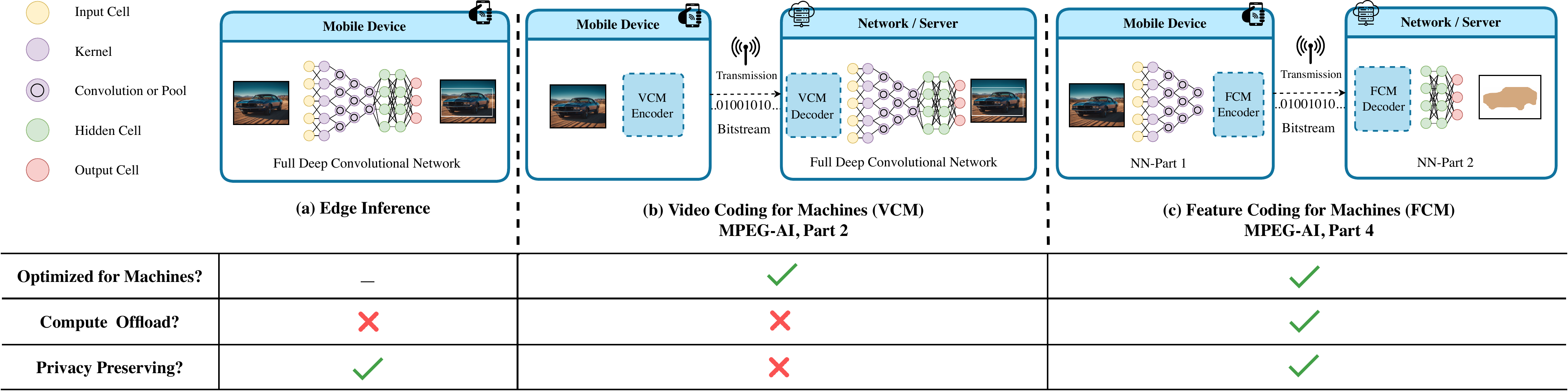}
  \caption{Overview of Coding for Machines (a) Edge Inference (b) Video Coding for Machines (VCM)  (c) Feature Coding for Machines (FCM)}
  \Description{Overview of Coding for Machines}
  \label{fig:teaser}
\end{teaserfigure}

\received{08 December 2025}

\maketitle

\begin{figure*}[t]
    \centering
    \begin{minipage}[b]{1\linewidth}
    \centering
    \includegraphics[width=\textwidth]{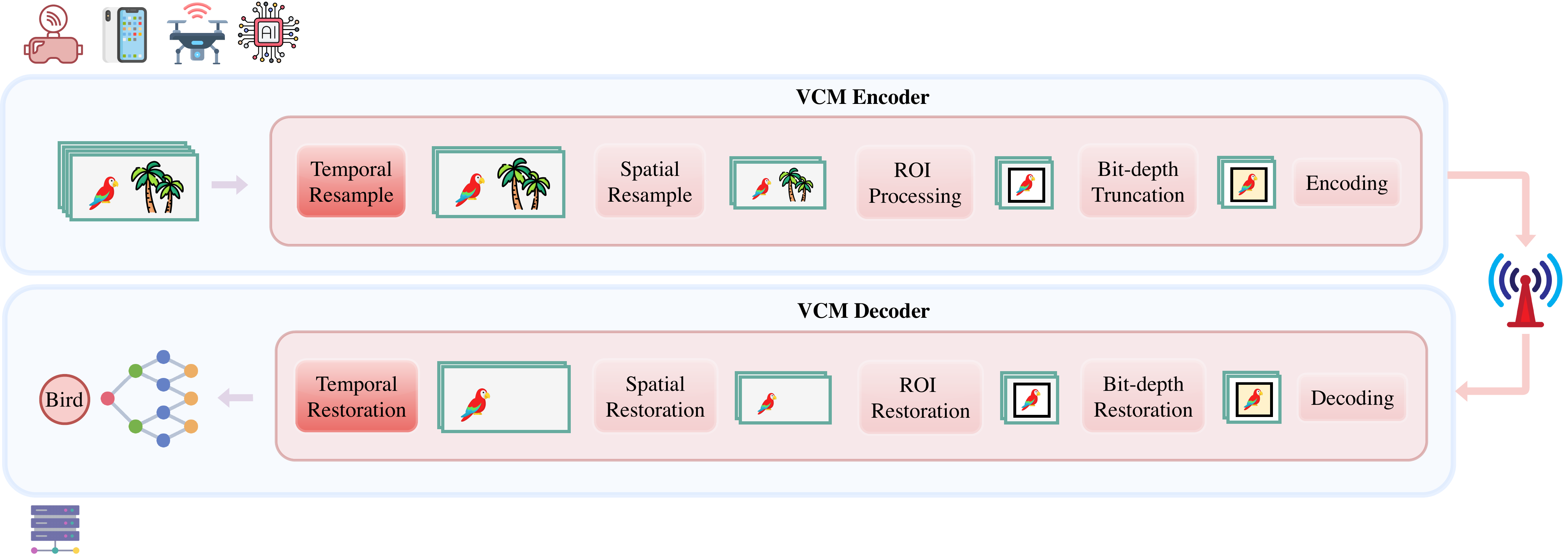}
    \end{minipage}
\caption{Overview of Video Coding for Machines (VCM)}
\label{fig:vcm_overview}
\end{figure*}

\section{Introduction}

In the current era of artificial intelligence (AI), machines have become primary consumers of visual data. Images captured by mobile phones, autonomous vehicles, surveillance systems, and embedded sensors are increasingly processed by AI-driven perception models to support a wide range of analytics. A direct approach for intelligent analytics is to execute the entire AI model on the device, as depicted in Figure~\ref{fig:teaser}~(a). This strategy is desirable because it protects privacy by keeping sensitive visual content local. However, achieving high analytic precision often requires state of the art deep learning models that exceed the computational and memory budgets of resource constrained platforms such as Internet of Things (IoT) sensors and smart cameras. As a result, many practical systems still rely on remote inference, where the edge device captures video and sends it to a cloud based model for downstream processing. Although this approach is simple and convenient, it does not scale well and depends on decades long pipelines  optimized for human perception rather than machine centric application~\cite{vcm_roi, vcm_acm_mm}. These limitations motivated the development of a new machine-to-machine video coding standard, Video Coding for Machines (VCM)~\cite{MPEG-AI-2},  as shown in Figure~\ref{fig:teaser}~(b). Moving Picture Experts Group (MPEG) initiated the development to create interoperable visual representations optimized for AI-based analysis tasks. 

VCM represents an important step toward machine-centric video coding. However, several challenges remain. VCM still transmits raw images to cloud servers, which leads to clear privacy concerns. In addition, when all computation is performed remotely, the increasing compute capability of modern edge devices remains underutilized. Although many edge platforms cannot run full deep networks, they are often equipped with neural processing units (NPUs) or lightweight AI accelerators that can support partial inference.

To address these limitations, a different architectural paradigm has gained momentum~\cite{fcm, fcm_iscas,fcm_demo, cem_fcm, fcm_jm}. Rather than transmitting full images to the cloud, the system executes part of the AI model on the device and sends only the intermediate representations, as depicted in Figure~\ref{fig:teaser}~(c). The device processes the early layers of the network and produces feature tensors that capture the semantic information required for the downstream task. Transmitting these features reduces bandwidth requirement, improves privacy, and enables compute offload. This paradigm is also attractive to AI service providers because shifting the initial stages of computation to edge devices can reduce cloud operating costs. At the same time, it benefits users by allowing state of the art models to run on existing hardware. As newer models require increased computation, users can compute the early layers locally and offload the remaining layers to the cloud. This architecture of split compute further enables collaborative intelligence~\cite{hm_collab, hm_back_forth}. In multi-drone systems, for example, a drone with limited battery capacity can request another drone to process part of the model. A drone can perform the early portion of inference and offload the remainder to one or more nearby drones with higher battery levels and computing power. This strategy allows the mission to continue without interruption while maintaining efficient use of shared resources.

Building on this idea, the nature of what is transmitted also changes. Instead of sending actual frames, the system communicates internal neural activations produced by the early layers of the model. The network is partitioned so that the device executes the front part of the architecture and the server completes the remaining layers. The transmitted representations are therefore task oriented and privacy preserving because they no longer resemble raw images. 

This shift, however, introduces a new compression problem. Intermediate features have statistics that differ substantially from natural images, which makes traditional video codecs ineffective~\cite{vcm_standard_cfe}. As a result, a dedicated solution is required to support this new style of machine-to-machine communication. To meet this need, MPEG initiated the development of Feature Coding for Machines (FCM) standard~\cite{MPEG-AI-4}, which defines coding tools optimized to compress intermediate feature tensors and enables collaborative intelligence across distributed devices.

This paper provides an overview of both VCM and FCM, followed by a comprehensive experimental comparison that evaluates VCM, FCM, and conventional remote inference schemes based on standard video codecs. In addition, we examine the impact of the inner codec within the FCM pipeline and analyze the performance of Advanced Video Coding (AVC)~\cite{avc}, High Efficiency Video Coding (HEVC)~\cite{hevc}, and Versatile Video Coding (VVC)~\cite{vvc} on downstream tasks. Our experiments show that although these codecs differ substantially in human centric visual quality metrics such as PSNR, their effect on many machine perception tasks remains surprisingly small. Finally, the paper concludes with a discussion of current limitations in VCM and FCM and outlines promising directions for future research.

\section{Video Coding for Machines}
\label{sec:vcm}
In this section, we provide a brief overview of the coding tools used in Video Coding for Machines (VCM)~\cite{descVCM}, also known as MPEG-AI Part 2. VCM adopts an H.26X family codec as its inner codec and augments it with a set of machine-centric tools that target task driven optimization, as illustrated in Figure~\ref{fig:vcm_overview}. 

Given an input video, the pipeline begins by reducing temporal redundancy through a temporal resampling module, which removes frames that do not introduce meaningful semantic change. Next, a spatial resampling module adapts the spatial resolution based on object occupancy and machine analytics requirements, retaining only the detail necessary for downstream tasks. Afterward, region based processing is applied through the region of interest processing module, which reallocates spatial resources toward semantically important regions while simplifying background content. Finally, before encoding, a bit-depth truncation module lowers the sample precision to reduce bandwidth with minimal impact on machine performance. A more detailed description of each coding tool is described below.

\vspace{0.5em}
\noindent\textbf{Temporal Resample.}
Temporal resampling reduces frame rate by identifying and discarding frames that do not contribute new semantic information. Inter-frame similarity is assessed either through object tracking or foreground based analysis. In the tracking based mode, similarity is measured by comparing bounding boxes of objects with consistent track IDs and computing their intersection over union (IoU). Frames with sufficient correspondence are grouped into a Group of Similar (GoS) pictures, from which only the first frame is retained. In the detection based mode, temporal sampling is adapted according to scene dynamics, where object displacement, coverage, and luminance driven differences determine the sampling interval. VCM also offers, a scalar ratio mode offers a deterministic setting that discards frames following fixed ratios of 2:1, 4:1, or 8:1. All decisions, including GoS configuration and the number of skipped frames, are explicitly signaled in the bitstream to enable reconstruction of the dropped frames at the decoder.

\vspace{0.5em}
\noindent\textbf{Spatial Resample.}
Spatial resampling selects an optimal resolution for each picture using Object Occupancy Distribution (OOD) computed across candidate scale factors. Downsampling is performed through VVC reference picture resampling filters, yielding adaptively scaled pictures that preserve object integrity for machine perception with reduced spatial redundancy.

\vspace{0.5em}
\noindent\textbf{ROI Processing.}
Region of Interest (ROI) processing enables VCM to focus coding resources on semantically relevant areas. For instance, in an image containing a bird and a tree, as illustrated in Figure~\ref{fig:vcm_overview}, only the bird region may be necessary for the downstream task, so background content can be removed to reduce bandwidth requirement. This ROI processing module begins with ROI extraction using the defined detection network, followed by margin-dilation to account for localization uncertainty. Spatially adjacent ROIs are merged based on horizontal and vertical proximity, and temporal stability is achieved by accumulating ROIs over a fixed window. 

\vspace{0.5em}
\noindent\textbf{Bit-depth Truncation.}
Bit-depth truncation reduces the sample precision of the luma and chroma components to a level sufficient for machine perception. The encoder applies right shifts to the sample values according to the configured truncation parameters, effectively discarding least significant bits that contribute little to downstream task accuracy. Since many machine vision models exhibit limited sensitivity to fine value quantization, this operation yields bitrate reduction with minimal impact on recognition performance. To preserve luminance characteristics when needed, optional enhancement modules can be applied, including tile-based histogram equalization and contrast normalization. 

\vspace{0.5em}
\noindent\textbf{Encoding \& Decoding.}
In this stage, the pre-processed frames are encoded using a conventional video codec such as AVC, HEVC, or VVC, while all VCM specific side information is embedded as metadata. 

\vspace{0.5em}
\noindent\textbf{Bit-depth Restoration.}
The VCM decoder first restores numerical precision by left shifting the luma and chroma samples according to the signaled parameters. 

\vspace{0.5em}
\noindent\textbf{ROI Restoration.}
ROI restoration reconstructs the spatial layout of the frame. Rows and columns removed during ROI retargeting are reinserted using the transmitted ROI grid. 

\vspace{0.5em}
\noindent\textbf{Spatial Restoration.}
Spatial restoration then upsamples the reconstructed frame to its target resolution using the reference picture resampling filters of VVC. 

\vspace{0.5em}
\noindent\textbf{Temporal Restoration.}
Finally, temporal restoration synthesizes the frames dropped during temporal resampling. Depending on whether adaptive or scalar ratio sampling was applied, a learned interpolation network estimates intermediate frames by refining motion fields and fusion masks across multiple hierarchy levels. 

Finally, the reconstructed video is fed to AI model to get predictions for the required task.

\begin{figure*}[t]
    \centering
    \begin{minipage}[b]{1\linewidth}
    \centering
    \includegraphics[width=\textwidth]{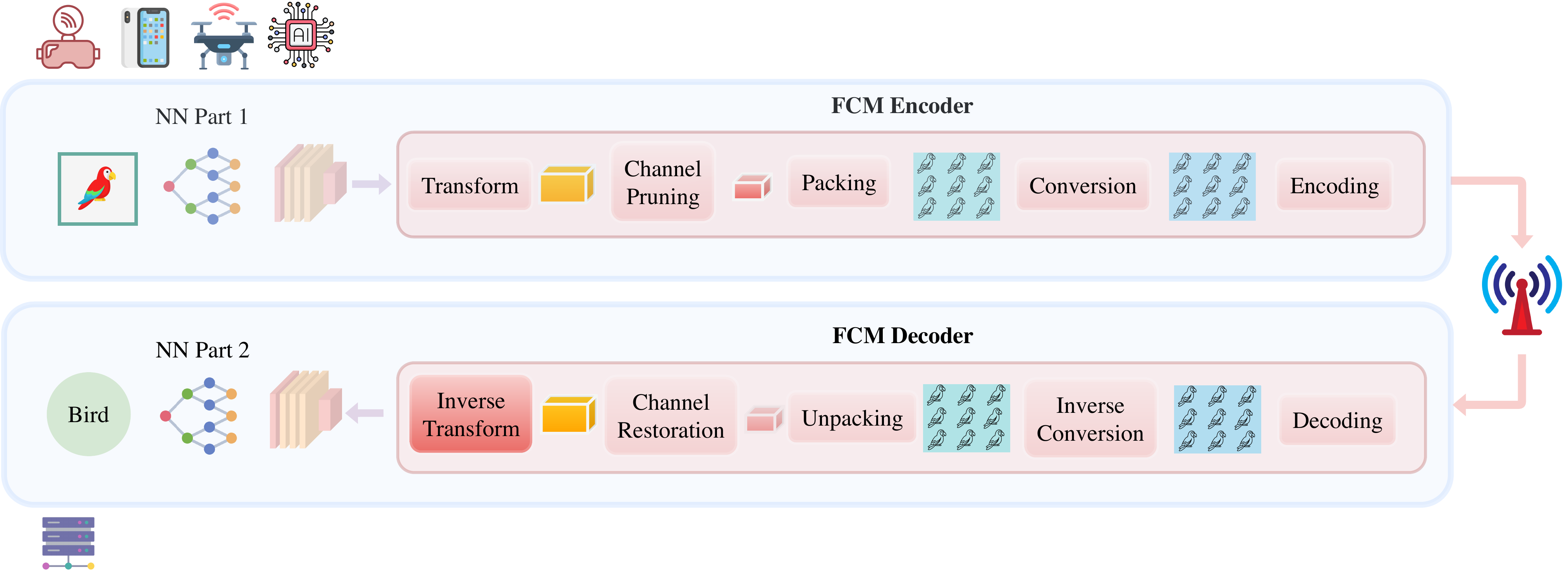}
    \end{minipage}
\caption{Overview of Feature Coding for Machines (FCM)}
\label{fig:fcm_overview}
\end{figure*}

\section{Feature Coding for Machines}
\begin{algorithm}[t]
\caption{LCR Encoding of Channels Indices}
\label{alg:lcr-encoding}
\begin{algorithmic}[1]
\State \textbf{Input:} Pruned channel indices $I = \{i_0, \ldots, i_{k-1}\}$, where $0 \le i_0 < \cdots < i_{k-1}$
\State \textbf{Output:} Number of pruned channels $k$ and combination index $\text{rank}$
\State $k \gets |I|$
\State $\text{rank} \gets 0$
\State $i_{-1} \gets -1$
\For{$t = 0$ to $k-1$}
    \For{$j = i_{t-1} + 1$ to $i_t - 1$}
        \State $\text{rank} \gets \text{rank} + \binom{N - j - 1}{k - t - 1}$
    \EndFor
\EndFor
\State \Return $(k, \text{rank})$
\end{algorithmic}
\end{algorithm}

Feature Coding for Machines (FCM) is designed to compress intermediate neural features that arise inside modern deep networks. The framework consists of a set of coding tools optimized to reduce, quantize, and restore feature representations while preserving end-task accuracy~\cite{descFCTM}. A typical FCM encoder and decoder structure is shown in Figure~\ref{fig:fcm_overview}. The coding tools are described below.

\vspace{0.5em}
\noindent\textbf{Transform.}
The input to the FCM encoder is the intermediate feature tensor produced by the first part of a split neural network. Let \(X_{t} \in \mathbb{R}^{C \times H \times W}\) denote the intermediate feature tensor at layer \(t\) in a multiscale feature pyramid. This feature tensor often has far higher dimensionality than the input image. The high-dimensionality is caused by the multiscale nature of model feature extractors as well as redundant correlations across channels.

The goal of the Transform module is to reduce the dimensionality of \(X_t\) by mapping it to a more compact representation while decorrelating its values as much as possible. FCM allows users to design any learned transform suited to their architecture and deployment constraints. Since the Transform step and downstream lossy coding can change the underlying feature distribution, FCM requires that the encoder transmit the global statistics of \(X_t\). These statistics are defined as
\begin{equation}
\mu = \mathbb{E}[X_{t}], \qquad
\sigma = \sqrt{\mathbb{E}\big[(X_t - \mu)^{2}\big]}
\label{eq:enc_stats_final_X}
\end{equation}
where $\mu$ and $\sigma$ are computed over all elements of \(X_{t}\). These parameters are necessary for the final refinement of the reconstructed feature. More details about this tool can be found in~\cite{fcm_iscas}.

\vspace{0.5em}
\noindent\textbf{Channel Pruning.}
Even after dimensionality reduction, many channels in the transformed representation may remain redundant. Depending on the input content, only a subset of channels is required for accurate inference. The Channel Pruning module removes less important channels to reduce bitrate. FCM uses the Lexicographic Combinatorial Rank (LCR)~\cite{eimran_lcr} representation to signal the indices of pruned channels. The encoding procedure is given in Algorithm~\ref{alg:lcr-encoding}.

\begin{algorithm}[t]
\caption{LCR Decoding of Channels Indices}
\label{alg:lcr-decoding}
\begin{algorithmic}[1]
\State \textbf{Input:} 
$k$ (number of pruned channels), 
$\text{rank}$ (lexicographic combination index),
$N$ (total number of channels)
\State \textbf{Output:} Pruned channel indices  $I = \{i_0, \ldots, i_{k-1}\}$
\State $\text{temp} \gets \text{rank}$
\State $x \gets 0$
\State $I \gets [\ ]$
\For{$t = 0$ to $k - 1$}
    \While{$\binom{N - x - 1}{k - t - 1} \le \text{temp}$}
        \State $\text{temp} \gets \text{temp} - \binom{N - x - 1}{k - t - 1}$
        \State $x \gets x + 1$
    \EndWhile
    \State $i_t \gets x$
    \State Append $i_t$ to $I$
    \State $x \gets x + 1$
\EndFor
\State \Return $I$
\end{algorithmic}
\end{algorithm}

\vspace{0.5em}
\noindent\textbf{Packing.}
After pruning, the reduced feature tensor is spatially arranged into a 2D frame suitable for compression using a conventional image or video codec. FCM optionally supports Channel Rearrangement and Channel Resizing to further lower entropy and improve compression efficiency.

\vspace{0.5em}
\noindent\textbf{Conversion.}
The packed feature frame contains 32-bit floating point values. To enable the use of conventional codec, these values are linearly mapped to n-bit (usually 10-bit) unsigned integers. Since deep learning networks are sensitive to global feature statistics~\cite{fcm_iscas}, the encoder computes the global mean and standard deviation of the reduced feature representation
\begin{equation}
\mu_{x} = \mathbb{E}[x], \qquad
\sigma_{x} = \sqrt{\mathbb{E}\big[(x - \mu_{x})^{2}\big]}
\label{eq:enc_reduced_stats_x}
\end{equation}
where \(x \in \mathbb{R}^{C \times H \times W}\). The pair \((\mu_{x}, \sigma_{x})\) is transmitted to ensure distribution correction after decoding.

\vspace{0.5em}
\noindent\textbf{Encoding \& Decoding.}
After conversion, the packed frame is encoded using a standard codec. All metadata is transmitted. At the decoder side, the packed frame is reconstructed and the metadata is extracted. The pruned channel indices are recovered using Algorithm~\ref{alg:lcr-decoding}. If Channel Resizing was used, its index information is also decoded with the same Algorithm~\ref{alg:lcr-decoding}.


\begin{figure}[t]
    \centering
    \begin{minipage}[b]{1\linewidth}
    \centering
    \includegraphics[width=\textwidth]{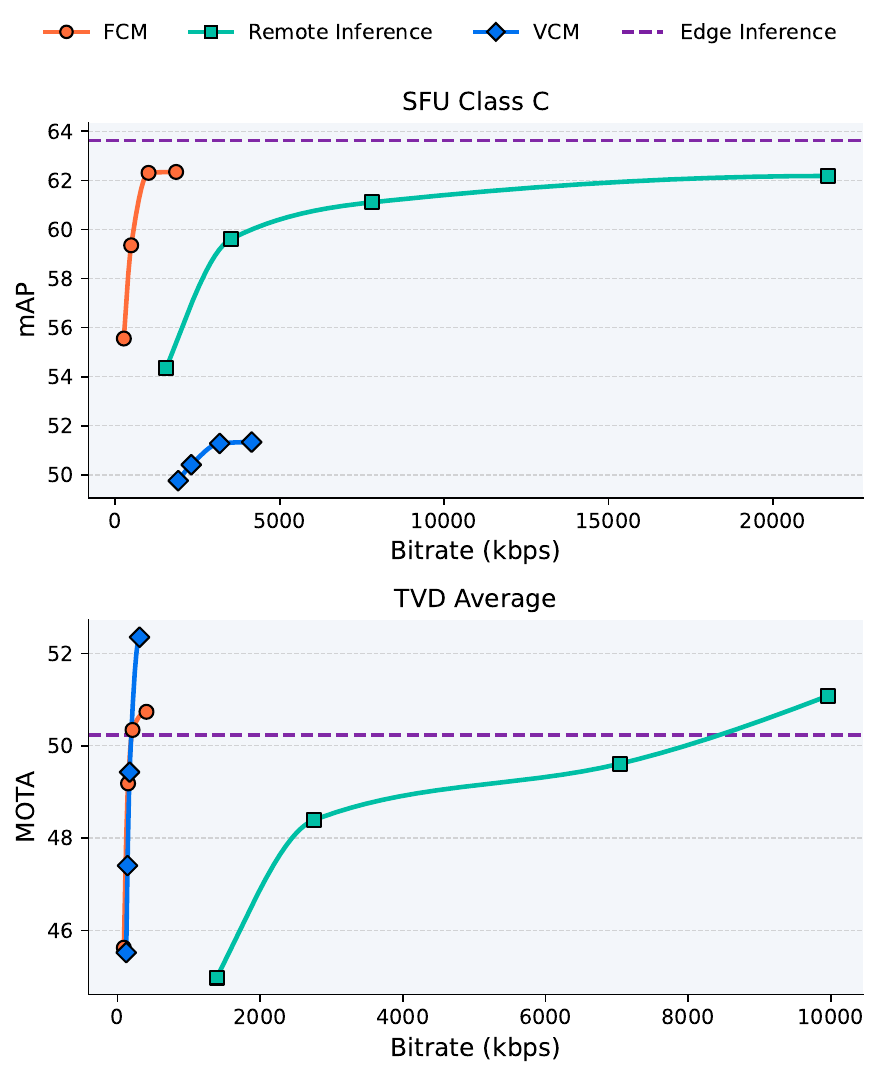}
    \end{minipage}
\caption{Rate-accuracy comparison of VCM, FCM, edge inference, and remote inference on SFU Class C and TVD Average, following the FCM Common Training and Test Conditions (CTTC).}

\label{fig:bd_rate_fcm_cttc}
\end{figure}

\vspace{0.5em}
\noindent\textbf{Inverse Conversion.}
The reconstructed packed frame is converted back to 32-bit floating point. The unrefined reduced feature is denoted by \(\hat{x}\). Its global statistics are computed as
\begin{equation}
\hat{\mu}_{\hat{x}} = \mathbb{E}[\hat{x}], \qquad
\hat{\sigma}_{\hat{x}} = \sqrt{\mathbb{E}\big[(\hat{x} - \hat{\mu}_{\hat{x}})^{2}\big]}
\label{eq:dec_reduced_stats_x}
\end{equation}
and refinement is performed using transmitted values calculated by the Equation~\ref{eq:enc_reduced_stats_x} 
\begin{equation}
\hat{x}^{refined} = \sigma_{x}\frac{\hat{x} - \hat{\mu}_{\hat{x}}}{\hat{\sigma}_{\hat{x}}} + \mu_{x}
\label{eq:refine_reduced_x}
\end{equation}
This step aligns the decoder distribution~\cite{fcm_iscas} with the encoder-side distribution and significantly improves downstream accuracy.

\vspace{0.5em}
\noindent\textbf{Unpacking.}
The refined packed frame is converted back to a 3D tensor with shape \(\hat{x}^{refined}\). This restores the spatial and channel structure of the reduced feature.

\vspace{0.5em}
\noindent\textbf{Channel Restoration.}
Channels that were removed during pruning are restored. Each pruned channel is filled with a scalar value equal to the mean of the current feature tensor to minimize distribution shift.

\vspace{0.5em}
\noindent\textbf{Inverse Transform.}
The restored reduced feature tensor is mapped back to the original multiscale feature space using the learned inverse transform. Let the unrefined reconstructed feature be \(\hat{X}_{t}\). The decoder computes its statistics as
\begin{equation}
\hat{\mu} = \mathbb{E}[\hat{X}_{t}], \qquad
\hat{\sigma} = \sqrt{\mathbb{E}\big[(\hat{X}_{t} - \hat{\mu})^{2}\big]}
\label{eq:dec_stats_final_X}
\end{equation}
and applies the final refinement using the encoder side parameters calculated by Equation~\ref{eq:enc_stats_final_X}
\begin{equation}
\hat{X}_t^{refined} = \sigma\frac{\hat{X}_{t} - \hat{\mu}}{\hat{\sigma}} + \mu
\label{eq:final_refine_X}
\end{equation}
This restores the global statistics~\cite{fcm_iscas} of the original intermediate feature tensor before feeding it into the second part of the network to complete the inference process.

\section{Evaluation}
\label{sec:evaluation}

\subsection{Experimental Setup}

To compare the coding efficiency of VCM and FCM, we follow the official common test conditions defined by MPEG. In both cases, the evaluation is restricted to the datasets, tasks, and network configurations that are common across the two standards, and all parameters are aligned accordingly. From the VCM CTC~\cite{vcm_cttc}, we use the random access (RA), low delay (LD), and all intra (AI) configurations together with the end to end (E2E) pipeline. From the FCM CTTC~\cite{fcm_cttc}, we use the LD configuration. In the FCM CTTC, an intra refresh frame is inserted periodically based on the frame rate of the input sequence, matching the intra period applied in the RA configuration of the VCM CTC. All experiments are conducted using the official CompressAI-Vision~\cite{choi2025compressaivisionopensourcesoftwareevaluate} framework. For FCM, we use the Feature Coding Test Model (FCTM)~\footnote{https://git.mpeg.expert/MPEG/Video/fcm/fctm}, and for VCM we use the Video Coding for Machines Reference Software (VCM-RS)~\footnote{https://git.mpeg.expert/MPEG/Video/VCM/VCM-RS}.

For the SFU dataset~\cite{sfu_v1} and the object detection task, we adopt the Faster R-CNN X101~\cite{faster_rcnn} network and follow the split defined in the FCM CTTC. The network is partitioned after the feature pyramid network, and the four multi scale feature maps P2 to P5 are extracted and processed by the FCM pipeline. In the remote inference baseline and in the VCM pipeline, the full Faster R-CNN model operates directly on the decoded frames. For the TVD dataset~\cite{tvd} and the tracking task, the split inference configuration follows the CTTC specification, where the Joint Detection and Embedding (JDE)~\cite{wang2019towards} model is partitioned at the split point near Darknet-53.


\subsection{Comparison of VCM and FCM}
Figure~\ref{fig:bd_rate_fcm_cttc} presents a comparison of rate accuracy performance for VCM and FCM.
For SFU Class C, the edge inference reference achieves an mAP of 63.71. FCM approaches edge inference while operating at very low bitrates. In the 62.30 mAP region, FCM requires only about 1850 to 1000 kbps, which corresponds to a reduction of approximately 90\% to 95\% compared to remote inference, which needs about 22000 to 7800 kbps to reach similar accuracy.  In contrast, VCM reaches only about 51.34 mAP at more than 4100 kbps. This means that VCM spends almost eight times more bitrate than low rate FCM to achieve noticeably lower accuracy. A similar trend appears in the TVD Average tracking results. The edge inference achieves a MOTA of 50.23. FCM reaches 50.73 MOTA at only 406 kbps, representing more than a 97\% bitrate reduction compared to remote inference, which requires about 9961 kbps for a comparable score. At around 148 kbps, FCM still maintains 49.18 MOTA, while remote inference needs about 2753 kbps to reach 48.39 MOTA. VCM performs well in the high bitrate region but its accuracy drops sharply below 300 kbps, whereas FCM degrades more steadily. Another interesting observation in the TVD tracking experiments is that both FCM and VCM both surpass the edge inference accuracy at high bitrates. For example, FCM reaches 50.73 MOTA and VCM reaches 52.35 MOTA, compared to 50.23 MOTA for edge inference. It is speculated that mild compression noise introduced by FCM tools help improving the performance of the end-task.

\begin{figure}[t]
    \centering
    \begin{minipage}[b]{1\linewidth}
    \centering
    \includegraphics[width=\textwidth]{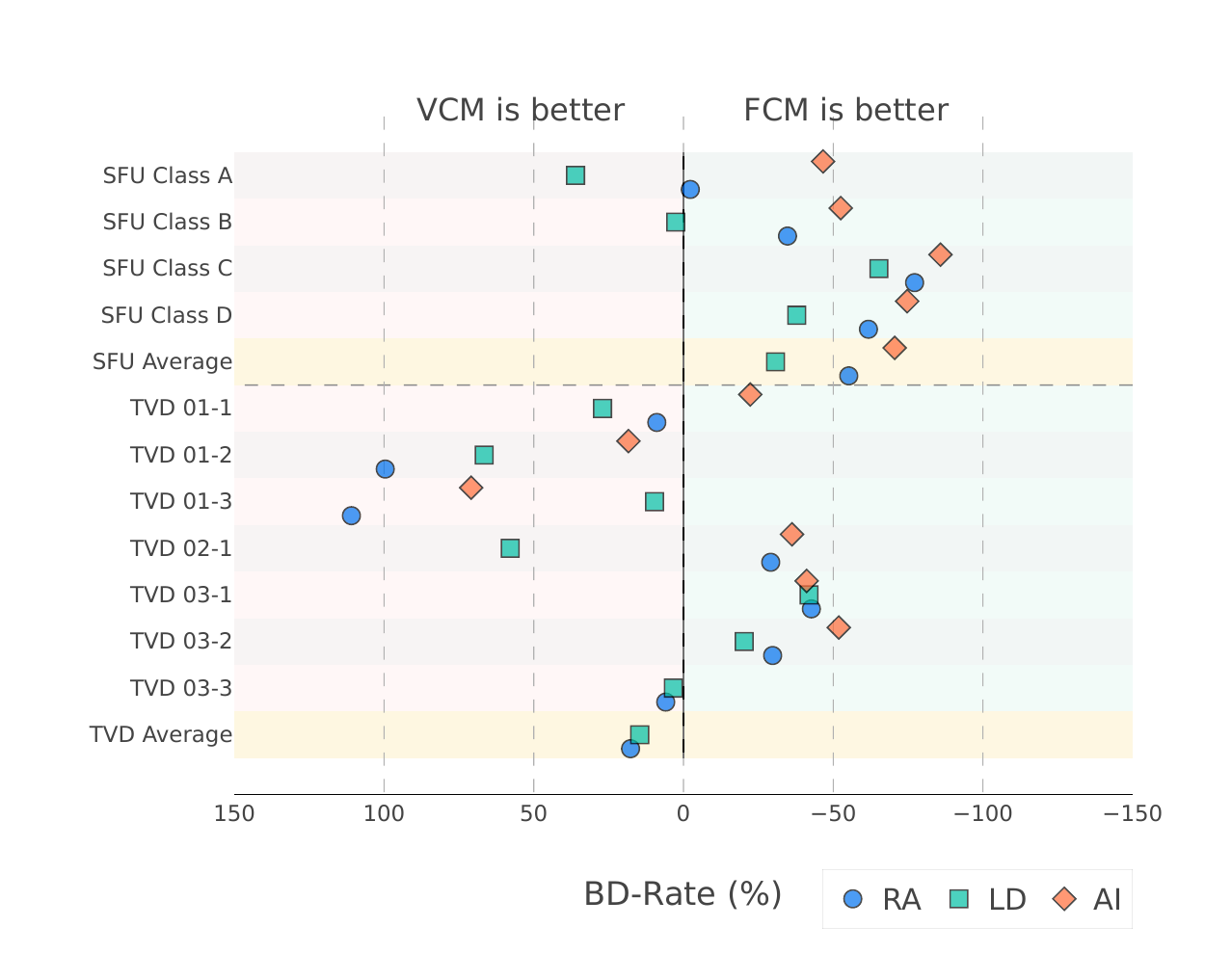}
    \end{minipage}
\caption{BD-Rate results of FCM with respect to the VCM anchor under the VCM Common Test Conditions (CTC). Results are shown for RA, LD, and AI configurations across SFU object detection sequences and TVD object tracking sequences. Negative BD-Rate indicates lower bitrate for equal task performance and positive BD-Rate indicates higher bitrate.}

\label{fig:bd_rate_vcm_cttc}
\end{figure}

Figure~\ref{fig:bd_rate_vcm_cttc} reports the BD-Rate of FCM with respect to the VCM anchor under the VCM CTC for three coding configurations (RA, LD, and AI) on both SFU object detection and TVD object tracking sequences. Negative BD-Rate values indicate that FCM achieves the same task performance at a lower bitrate than VCM, while positive values indicate that VCM is more efficient.

For the SFU detection sequences, FCM consistently outperforms VCM. Under RA, the BD-Rate ranges from \(-2.28\%\) on Class~A to \(-77.20\%\) on Class~C, with an average of \(-55.21\%\), which shows that FCM reduces bitrate by more than half on average compared to VCM for the same mAP. The LD configuration exhibits a similar trend, with large gains on Class~C and Class~D (\(-65.30\%\) and \(-37.85\%\)) and a modest loss on Class~A, leading to an average BD-Rate of \(-30.73\%\). The AI configuration provides the strongest savings, with BD-Rate values between \(-46.63\%\) and \(-85.84\%\) across all classes and an average of \(-70.54\%\). 

The TVD tracking results are more mixed. For RA, BD-Rate values range from large positive values on sequences TVD~01-2 and TVD~01-3 (\(+99.62\%\) and \(+110.92\%\)) to significant negative values on TVD~02-1, TVD~03-1, and TVD~03-2 (\(-29.14\%\), \(-42.72\%\), and \(-29.78\%\)), which yields an average of \(+17.68\%\). LD shows a similar pattern, with strong gains on some sequences and losses on others, resulting in an average BD-Rate of \(+14.61\%\). 

From these observations in Figure~\ref{fig:bd_rate_fcm_cttc} and Figure~\ref{fig:bd_rate_vcm_cttc}, one could conclude that both VCM and FCM achieve better performance on the tracking task compared to traditional remote inference. For detection, FCM delivers overall strong performance, while VCM sometimes falls short.

\begin{table}[t]
\centering
\caption{BD-Rate results of HEVC (HM-18.0) and AVC (JM-19.1) with respect to the VVC anchor (VTM 23.3) across object detection (SFU) and object tracking (TVD Average).}
\renewcommand{\arraystretch}{1.2}
\setlength{\tabcolsep}{10pt}
\begin{tabular}{lccc}
\toprule
\textbf{Dataset} & \textbf{Task} & \textbf{HEVC} & \textbf{AVC} \\
\midrule
SFU Class A/B & Detection & 3.92\%  & 29.03\% \\
SFU Class C   & Detection & 2.69\%  & 50.59\% \\
SFU Class D   & Detection & 0.76\%  & 40.70\% \\
TVD Average   & Tracking  & -1.81\% & 8.79\%  \\
\midrule
\textbf{Average} &  & \textbf{1.39\%} & \textbf{32.28\%} \\
\bottomrule
\end{tabular}
\label{tbl:bd_rate_avc_hevc}
\end{table}

\subsection{Impact of H.26X Codec in FCM}
Traditional video codecs are designed to minimize distortion in the pixel domain, with human visual quality as the primary optimization target. In contrast, FCM compresses intermediate neural features that have been strongly decorrelated by deep network transformations and exhibit temporal behavior that differs significantly from natural video signals. As a consequence, many advanced coding tools that provide large PSNR gains in modern codecs do not necessarily produce proportional improvements in downstream machine performance. To examine this behavior, we compare the effect of different inner codecs within FCM, specifically replacing the VVC anchor with HEVC and AVC. The BD-Rate results in Table~\ref{tbl:bd_rate_avc_hevc} show that HEVC maintains task accuracy very close to VVC. AVC, in contrast, introduces a higher average BD-Rate of 32.28\%. 

This has important implications for deployment. HEVC is widely supported in existing hardware and software ecosystems, and many commercial devices already include dedicated HEVC accelerators. When the machine task performance difference relative to VVC is small, as observed here, an FCM system can operate effectively using HEVC without requiring immediate migration to VVC. This makes HEVC an attractive option for near term FCM deployment, particularly in scenarios where codec availability, computational cost, or licensing constraints limit the adoption of VVC.

\section{Conclusion and Future Direction}

In conclusion, both Video Coding for Machines (VCM) and Feature Coding for Machines (FCM) offer advantages over traditional remote inference for machine-to-machine communication. VCM provides coding tools optimized for pixel domain machine-to-machine communication, and FCM provides coding tools to compress intermediate neural features to preserve privacy and enable collaborative intelligence. Our analysis shows that in FCM the choice of inner codec has minimal impact on downstream performance, indicating that existing hardware can be used effectively for feature domain compression.

Future research should pursue near lossless task performance, aiming for accuracy drops within 1\% relative to edge inference rather than primarily targeting higher bitrate reduction, since even 1\% decrease in machine task accuracy can be significant in certain applications.

\bibliographystyle{ACM-Reference-Format}
\bibliography{ref}

\end{document}